\documentclass{article}
\usepackage[margin=2.5cm]{geometry}
\usepackage{graphicx}
\usepackage{amsmath}
\usepackage{amssymb}
\usepackage{braket}
\usepackage{subcaption}
\usepackage{xcolor}
\usepackage{booktabs}
\usepackage{hyperref}
\usepackage{authblk}
\usepackage[numbers,sort&compress]{natbib}

\begin{document}

\title{A path to natural language through tokenisation and transformers}

\author[1,a]{David S. Berman}
\author[1,b]{Alexander G. Stapleton}

\affil[1]{Centre for Theoretical Physics, Queen Mary University of London, Mile End Road, London E1\,4NS, United Kingdom}

\affil[a]{\texttt{a.g.stapleton@qmul.ac.uk}}
\affil[b]{\texttt{d.s.berman@qmul.ac.uk}} % Correspondence email for Marc S. Klinger

\maketitle

\begin{abstract}
Natural languages exhibit striking regularities in their statistical structure, including notably the emergence of Zipf's and Heaps' laws. Despite this, it remains broadly unclear how these properties relate to the modern tokenisation schemes used in contemporary transformer models. In this note, we analyse the information content (as measured by the Shannon entropy) of various corpora under the assumption of a Zipfian frequency distribution, and derive a closed-form expression for the slot entropy expectation value. We then empirically investigate how byte--pair encoding (BPE) transforms corpus statistics, showing that recursive applications of BPE drive token frequencies toward a Zipfian power law while inducing a characteristic growth pattern in empirical entropy. Utilizing the ability of transformers to learn context dependent token probability distributions, we train language models on corpora tokenised at varying BPE depths, revealing that the model predictive entropies increasingly agree with Zipf-derived predictions as the BPE depth increases. Attention-based diagnostics further indicate that deeper tokenisation reduces local token dependencies, bringing the empirical distribution closer to the weakly dependent (near IID) regime. Together, these results clarify how BPE acts not only as a compression mechanism but also as a statistical transform that reconstructs key informational properties of natural language.
\end{abstract}

\section{Introduction}

Transformers are the basis for the current success of generative language models. Their ability to auto-regressively generate text arises from having learnt the conditional probability distribution for a token conditioned on the previous tokens in a sequence. By obtaining this conditional probability distribution, a model will have implicitly learnt about the context dependent information content of natural language.

Studying the information content of language in this way is not novel; examining the information content of language goes back to Shannon who applied thermodynamic ideas, notably the concept of entropy, to communication channels. Shannon used the definition of entropy in terms of the probability of occurrence of the various tokens. This approach was seminal in how to interpret and calculate information content in signal processing. In practice though, true joint token probabilities are unknown, only the approximation where one assumes the tokens are independently distributed is easy to measure. This has been the approach to studying the information context of language for many years. Now, with the advent of large language models, we have the opportunity to train a model to learn the full conditional distribution of tokens and employ that to describe and investigate the information in a corpus. We anticipate that the general use of large language models to study the statistical nature of language will become more commonplace as scholars come to appreciate that these models have not just learnt to predict the {\it{next token}}, but the full probability distribution of tokens. This may have impact in signal processing, cryptography, and beyond, where the true statistical properties of language both matter, and may uncovered by the use of large language models.

In this work we wish to study the relationship between a well known statistical property of language, Zipf's law, and the depth of tokenisation in a language model. Before going into the details of this note, it is worth first commenting that there has been something of a lacuna in the study of the tokenisation process and its effect on the final performance of the LLM. The depth of tokenisation in most LLMs gets determined empirically as a hyperparameter choice. From a more abstract perspective, the tokenisation level is like the scale at which one examines language, with the primitive letter scale token the most fine grained. As we move to higher depths of tokenisation, the vocabulary grows as sub-word, word, or potentially multi-word and sentence level tokens emerge. This choice is analogous to the so called {\it{renormalisation}} scale in physics. Statistical field theories are described at a particular length scale which dictates their statistical properties. The so called {\it{renormalisation group (RG) flow}} describes how these properties change as one alters the scale, which amounts to inducing a flow in the model parameters. (For a link between learning in statistical models and exact renormalisation group flow see \cite{Berman_2023, Berman_2024}.) A key result in such theories is the so called c-theorem which states how the entropy in such a model changes as one undergoes an RG flow. Inspired by this phenomenon in physics, we will ask the equivalent question in language: ``how does the entropy in a corpus change as we alter the depth of tokenisation?"

Transformers offer a natural statistical lens to answer this question as the softmax provides a bona-fide probability distribution to study. We evaluate the entropy using this distribution, then calculate analytically the expected entropy of a text under the assumption that it obeys Zipf's law. Comparing these results allows us to demonstrate that deeper tokenisation leads to the predicted empirical entropy (as given by the transformer) getting closer and closer to the analytic expression for the entropy of a Zipfian distribution.

% The paper is structured as follows: we first introduce Zipf's law followed by transformers and byte-pair encoding. We then derive the analytic entropy of a Zipfian Corpus and comment on the most robust way to fit a Zipfian distribution to real data. (In principle, one needs just a single token frequency to fit to Zipf's law, but we demonstrate that fitting to mid-rank tokens provides the most stable way to empirically fit to data).
% With all this in place, we measure the entropy as a function of tokenisation depth, and compare to the predictions from Zipf's law. We end with some investigations of a scale in the text revealed by the attention mechanism that explains the path towards Zipfian statistics.

This note is organised as follows. We begin with an introduction to Zipf’s law, followed by background on transformer models and byte-pair encoding. We then derive the analytical entropy of a Zipfian corpus and discuss robust methods for fitting Zipfian distributions to empirical data. Although, in principle, a single token frequency suffices to determine a Zipfian fit, we show that restricting the fit to mid-rank tokens yields significantly greater empirical stability.

Building on this framework, we measure entropy as a function of tokenization depth and compare the results with theoretical predictions from Zipf’s law. Finally, we investigate a characteristic scale in text revealed by the attention mechanism, which provides insight into the emergence of Zipfian statistics.

\subsection{A primer of Zipf's law}
In general, almost all natural languages share a remarkable feature: the distribution of the ranks of words in a sufficiently large textual corpus approximately obey Zipf's law \cite{KingsleyZipf1932}. Zipf's law, named after linguist George Kingsley Zipf\footnote{Despite Zipf's observation of the power-law behaviour in the 1930s \cite{KingsleyZipf1932}, `Zipf's law' may actually be traced back to Estoup \cite{estoup1916}, who published his finding in 1916.}, states that the frequency of a word's occurrence $f_s$ is inversely proportional to its rank $r_s$,
\begin{equation}
	\label{eqn:zipf}
	f_s \sim \frac{1}{r_s},
\end{equation}
where the rank of a word refers to its position in an ordered largest-to-smallest frequency table. Despite its age, Zipf's law remains an active object of research, with modern studies continuing to examine the power law behaviour in modern applications, such as neural networks \cite{he2025pretrainedmodelsperformbest,berman2025zipfdistributionstwostagesymbolic}. Zipf's law emerges in almost all languages \cite{yu2018zipfslaw50languages}, where a number of works extend empirical observations of the power law behaviour \cite{berman2025morphemicoriginzipfslaw,berman2025zipfdistributionstwostagesymbolic} beyond human discourse, and even to the communications of animals (see e.g. \cite{doi:10.1126/sciadv.ads6014} which observes Zipf's law in whale communications). Whilst a rigorous derivation of Zipf's law has yet to be produced, it has been shown that one may derive a generalisation (the Mandelbrot-Zipf law \cite{mandelbrot1965information}) from the statistics of large corpora if each $n$-gram is equiprobable \cite{165464} -- of course, such a simplifying assumption is not valid in English (for example, the letter `e' appears approximately 170 times more than the letter `z' \cite{Lewand2000-ix}).

\begin{figure}[h]
    \centering
    \begin{subfigure}{0.48\linewidth}
        \centering
\raisebox{1.5ex}{%
            \includegraphics[width=\linewidth]{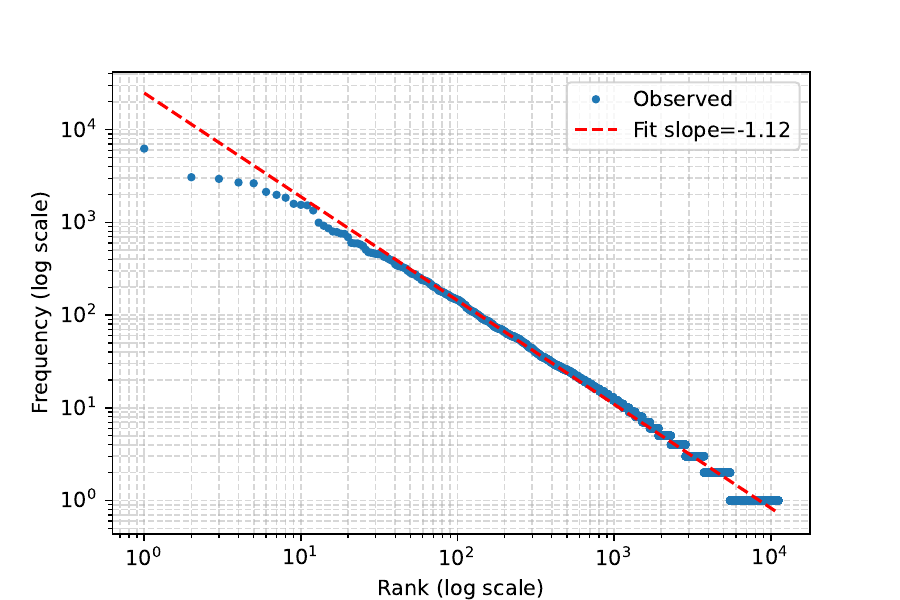}
        }
        \caption{Zipfian behaviour of subset of IMDB dataset}
        \label{fig:zipf_vanilla_imdb_subset}
    \end{subfigure}\hfill
    \begin{subfigure}{0.48\linewidth}
        \centering
        \includegraphics[width=\linewidth]{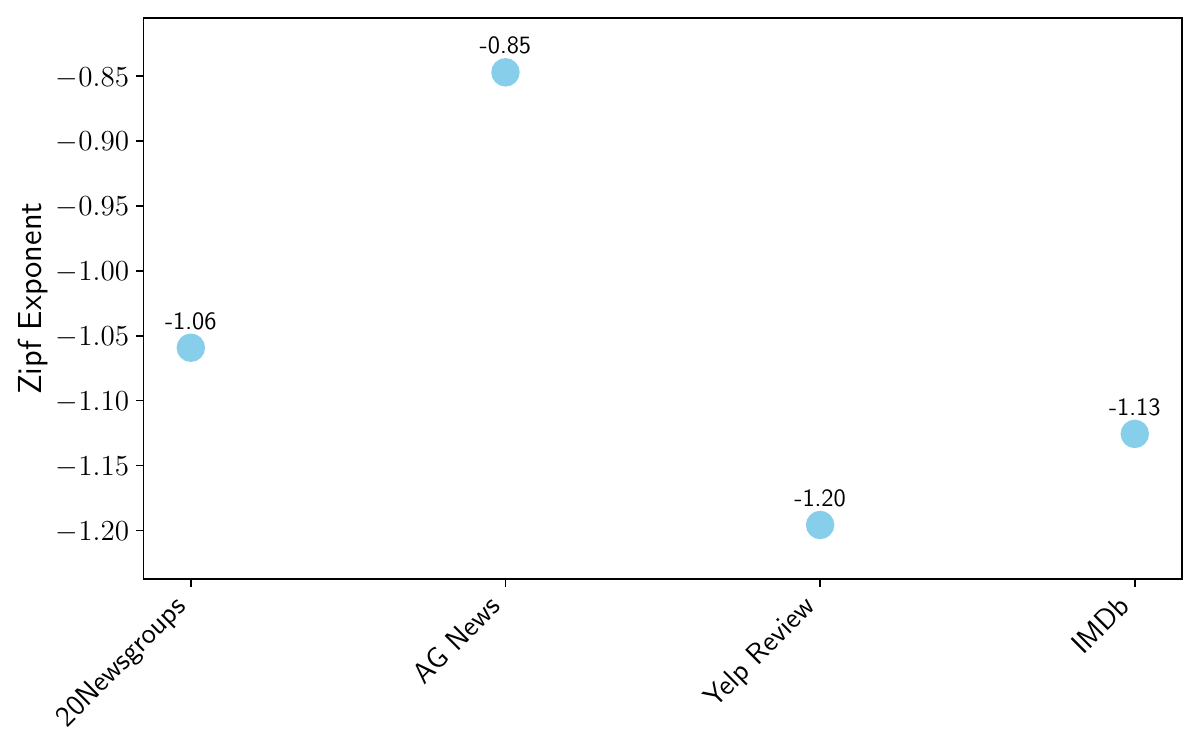}
        \caption{Fitted Zipf exponents for subsets of four corpora}
        \label{fig:zipf_exponents_four_datasets}
    \end{subfigure}
    \caption{Zipf statistics across datasets.}
\end{figure}

Whilst there exist many generalisations of Zipf's law amounting to higher-order corrections \footnote{Arguably the most notable of which is the aforementioned result of Mandelbrot, which states $f_s \sim \frac{1}{(r_s - a)^b}$ (where $a, b \in \mathbb R$ are determined by data).}, the most basic statement remains a remarkably good approximation for many corpora\footnote{Throughout this work, we treat Zipf’s law in its idealised form with exponent $\alpha=-1$. Empirical deviations are treated as subleading corrections.} (see, for example Figure \ref{fig:zipf_exponents_four_datasets}). Figure \ref{fig:zipf_vanilla_imdb_subset} shows explicitly the complete set of frequencies against ranks, whilst Figure \ref{fig:zipf_exponents_four_datasets} shows the fitted Zipf exponent for a selection of five common datasets.

\subsection{Encoding in transformers}
\label{sub:encoding_in_transformers}
One of the most impactful advances in the history of machine learning and human-machine interactions is arguably the transformer architecture. 

Transformers are a class of neural network architecture particularly suited to sequence processing tasks \cite{kämäräinen2025introductionsequencemodelingtransformers,vaswani2023attentionneed,turner2024introductiontransformers}, enabling effective handling of data through so-called self-attention mechanisms. Introduced in \cite{vaswani2023attentionneed}, transformers replace recurrent architectures and allow for parallel processing of input tokens, typically providing sample quality and model efficacy far beyond the capabilities of predecessor architectures. The ability of a transformer to encode long-range dependencies and context-dependent features has led to their almost universal adoption in an array of applications, including search engines and \textit{essentially all} modern large language models e.g.\cite{islam2023comprehensivesurveyapplicationstransformers, deepseekai2025deepseekv32pushingfrontieropen, chatgpt}.

A general transformer language model operates on sequences of discrete tokens, but the model itself processes continuous vectors. Therefore, an encoding scheme is required to map tokens into vector representations that the model can manipulate. Of course, one is free to choose (almost) any scheme to encode the sequence of $n$-grams, although a natural question is how one should optimise the scheme to yield desirable characteristics such as better model performance, faster loss convergence, etc. It has been found that transformer-based models typically perform better when the tokenisation scheme obeys Zipf's law \cite{he2025pretrainedmodelsperformbest}. One of the most common families of encodings are known as \textit{subword encodings} \cite{huggingfacecourse}. Of these, arguably the most ubiquitous is known as \textit{byte-pair encoding} \cite{gage1994new}, which provides a balanced vocabulary and avoids both the explosion of word-level vocabularies, and the inefficiency of character-level models. These encodings enable transformers to learn rich semantic relationships while preserving the sequential structure of language.

The byte-pair encoding (BPE) tokenization algorithm iteratively
merges the most frequent adjacent symbol pairs to construct a more compact, and
thus potentially more efficient, vocabulary. Let the corpus be represented as a
collection of \(M\) sequences of canonically ordered \footnote{Here canonical ordering refers to the ordering of tokens as spoken/written in natural language.}tokens,
\[
\mathcal D := \{ m_1, \ldots, m_M \},
\]
where each \(m_i\) represents a series of tokens constituting a message. The unique $n$-grams in the messages form a vocabulary $V^\iota = \{ s_i^\iota\}_{i=0}^{V^\iota}$ at the BPE recursion depth $\iota$. 

At each iteration, the algorithm identifies the most frequent adjacent pair of
n-grams,
\[
(\tilde \mu,\tilde\nu) = \arg\max_{(\mu, \nu)} \mathrm{freq}(s_\mu^\iota s_\nu^\iota),
\]
where \(\mathrm{freq}(s_\mu^\iota s_\nu^\iota)\) denotes the number of occurrences of the concatenated \textit{adjacent}
pairs $(s_\mu^\iota s_\nu^\iota)$ across the corpus. For computational efficiency, in practice, rather than greedily forming all pairs and checking frequency, one of course only checks pairs selected from those which occur in the dataset at the beginning of the iteration.

The most common pair is then merged into a new symbol $\tilde s^\iota$, and every
occurrence in each sequence is replaced according to
\[
m_i \leftarrow \mathrm{merge}_{s_{\tilde \mu} s_{\tilde \nu}}(m_i), \qquad i = 1,\ldots,M.
\]

After \(\iota\) iterations, the resulting vocabulary consists of the original $V^0$ tokens, as well as the $\iota$ newly introduced tokens. The resulting
tokenization scheme captures frequent subword patterns while remaining robust
to rare or previously unseen sequences.

From this procedure, we see that a BPE scheme may be generalised somewhat: instead of performing a single merge at each iteration, one may perform many. In this case, the algorithm is governed by two parameters:
the number of merges\footnote{Mnemonically, one should associated $\phi$ with the number of \textit{formations}.} \(\phi\) and the number of iterations, or recursions,  \(\iota\).
In all experiments where BPE is applied, we fix the number of merges to unity,
i.e. \(\phi = 1\), so each new iteration forms a single new token.

\section{Entropy of Zipfian Corpora}

\subsection{Theoretically predicted entropy of a Zipfian corpora}%
\label{sub:entropy_of_a_zipfian_corpus}

In this section we examine what restrictions the observation of Zipf's law place on the entropy of a statistical model. One considers a general statement of Shannon's famous relation $S = - \sum_i p_i \log p_i$, where $p_i$ is the probability of an event $\text{Prob}(X = x_i)$. Let $\mathcal D=\{m^\mu\}_{\mu=1}^M$ be a dataset consisting of $M$ draws from a distribution of messages, where each message $m^\mu$ contains $N^\mu$ $n$-grams. The set of $n$-grams constitute a vocabulary $V$. Assuming all messages are \textit{independent and identically distributed (IID)}, the probability of realising an $n$-gram of type $s$ can be empirically approximated by, in the large $\Gamma$ limit,
\begin{equation}
	\label{eqn:empirical_prob} 
	p_s = \frac{\sum_{\mu=1}^M n^\mu_s}{\Gamma},
\end{equation}
where $\Gamma := \sum_{\mu=1}^M N^\mu$ is the total number of $n$-grams in the dataset, and $n^\mu_s$ is the number of $n$-grams of type $s$ in the $\mu$-th message.

Furthermore, one may consider each message $m^\mu$ as a set of $N^\mu$ slots, each populated by an $n$-gram with a given probability. There is also an intrinsic probability associated to the length of each message (e.g. $N^\mu$), however in the subsequent calculations we always assume it has been realised a-priori.

In the most general setting, prior to any IID assumptions, the entropy of the probability distribution governing position $i$ of message $m^\mu$ is defined by
\begin{equation}
\label{eqn:IID_entropy}
	S^\mu_i := - \sum_{s=1}^{V} p_{s,i}^\mu \ln(p_{s,i}^\mu),
\end{equation}
where $p^\mu_{s,i} := \text{Prob}(t^\mu_i = s)$ is the probability that the $n$-gram\footnote{As a mnemonic, one may identify the $n$-gram $t_i^\mu$ as the $i$-th \textit{token} in the $\mu$-th message.} $t_{i}^\mu$ at position $i$ in message $m^\mu$ is of type $s$. One may thus calculate the average slot entropy across all messages and all slots per message, which, due to the IID condition, is given by
\begin{equation}
	S_* = \frac{1}{M} \sum_{\mu=1}^M \sum_{i=1}^{N^\mu} \frac{1}{N^\mu} S^\mu_i = S_0^0.
\end{equation}
In this case, $S_*$ is akin to the entropy density as it is an intensive quantity.

Suppressing the position and message indices (which convey no extra information as each slot is IID), the average entropy of each slot's token distribution may be written as,
\begin{align}
	\mathbb E[S] &= -\sum_{s=1}^V \frac{n_s}{\Gamma} \ln (n_s / \Gamma)\label{eqn:finite_entropy}\\ 
							 &= \frac{1}{\Gamma} \sum_{s=1}^V n_s (\ln \Gamma-\ln n_s) \label{eqn:S_in_terms_of_n},
\end{align}
where in the first line \eqref{eqn:empirical_prob} is used to approximate $p_s$.

If the distribution of the $n$-grams obey \eqref{eqn:zipf}, one may write $n_s = \frac{\gamma}{r_{s'}}$, where the index $s' =\rho(s)$, with $\rho(s)$ a ranking function which gives the position of the $s$-th $n$-gram in a ranked list of frequencies from largest to smallest\footnote{In order for further calculations to be mathematically sound, we impose that $\rho$ must be injective and surjective (i.e. a bijective function).}, and $\gamma$ a constant of proportionality intrinsic to the dataset $\mathcal D$. It is helpful to define the ordered frequency $f_s = n_{\rho(s)}$ such that $f_1 > f_2 > \ldots f_N$. By Zipf's law, we notice, albeit trivially, the remarkable fact that
\begin{equation}
	\label{eqn:zipf_recurse} 
	\frac{f_s}{f_{s+1}} = \frac{s+1}{s} \implies f_{s+1} = \frac{s}{s+1} f_s,
\end{equation}
where a key corollary of \eqref{eqn:zipf_recurse} arises from repeated applications:
\begin{equation}
\label{eqn:zipf_freq_identity}
f_s = \frac{f_1}{s}.
\end{equation}
Considering the inner sum of the second term in \eqref{eqn:S_in_terms_of_n}, since $\rho$ is bijective and the sum commutes, one may write
\begin{align}
	\sum_{s=1}^V f_s \ln f_s &= f_1 \ln f_1 + f_2 \ln f_2 + \ldots + f_V \ln f_V\\
											 &= f_1 (\ln f_1 + \frac{1}{2} \ln f_2 + \ldots + \frac{1}{V} \ln f_V)\\
											 &= f_1 (\ln f_1 + \frac{1}{2} \ln \frac{f_1}{2} + \ldots + \frac{1}{V} \ln \frac{f_1}{V})\\
											 &= f_1 (\ln f_1 + \frac{1}{2} (\ln f_1 - \ln 2) + \ldots + \frac{1}{V} (\ln f_1 - \ln V))\\
											 &= f_1 \ln f_1 (1 + \frac{1}{2} + \ldots + \frac{1}{V} ) -f_1 (\frac{1}{2} \ln 2 + \ldots + \frac{1}{V} \ln V) \\
											 &= f_1 \ln f_1 H_V + f_1 \alpha(V)
											 % &=: \phi_1(f; V),
\end{align}
where $H_V$ is the $V$-th harmonic number, and $\alpha(V) := -\sum_{s=2}^V \frac{\ln s}{s}$. Returning to the second term in  \eqref{eqn:S_in_terms_of_n}, one similarly finds,
\begin{equation}
	\sum_{s=1}^{V} f_s \ln \Gamma = \ln \Gamma f_1(1 + \frac{1}{2} + \ldots \frac{1}{V}) = H_V f_1 \ln \Gamma,
\end{equation}
where the second equality is realised by once more noting $f_s = f_1 f_s/f_1 = f_1/s$.

Returning to the entropy calculation, we arrive at the delightful expression
\begin{align}
\label{eqn:avg_entopy_zipf}
	\mathbb E[S] = \frac{f_1}{\Gamma}[H_V \ln \Gamma - \ln f_1 H_V - \alpha(V)],
\end{align}
depending only on a single $n$-gram frequency common to the whole dataset; the size of the vocabulary $V$; and the total number of $n$-grams in the corpus. We note that the entropy remains finite when the vocabulary size $V$ is finite; in the limit $V\to \infty$ the entropy diverges logarithmically for Zipf exponents $\alpha \leq 1$. Of course, in the previous calculation an implicit choice has been made in writing \eqref{eqn:avg_entopy_zipf} in terms of $f_1$. Using the Zipf identity \eqref{eqn:zipf_freq_identity},  \eqref{eqn:avg_entopy_zipf} may be written more generally as
\begin{align}
\label{eqn:avg_entopy_zipf_general}
	\mathbb E[S] = \frac{sf_s}{\Gamma}[H_V \ln \Gamma - \ln (sf_s) H_V - \alpha(V)].
\end{align}

\subsection{Numerical demonstration}
In this subsection we consider a series of numerical demonstrations to examine the empirical entropy and predicted entropies of the corpus, evaluated with  \eqref{eqn:avg_entopy_zipf}. 

\subsubsection{Dependence of the entropy calculation on $s$}
\label{ssub:dep_of_entropy_on_s}

As is apparent from  \eqref{eqn:avg_entopy_zipf_general}, one needs to establish (empirically fit) the frequency $f_s$ for the rank $s$ token. For simplicity in the above derviation $f_1$ was chosen. In fact,  it is intuitive to pick the position $s$ to be somewhere around the centre of the rank table as Zipf's law does not hold as well for extreme rankings. In the high-frequency end, the very top few words (such as common verbs) often occur more frequently than the idealised power-law curve predicts; the general linguistic structure of English forces them into heavy overuse \cite{baayen2001word}. Moreover, at the low-frequency end, rare words do not follow the expected smooth decay because of sampling effects, domain specific behaviour, and the presence of hapax legomena — words that appear only once in a single dataset \cite{baayen2001word}.

\begin{figure}[h]
    \centering
    \begin{subfigure}{0.45\linewidth}
        \includegraphics[width=\linewidth]{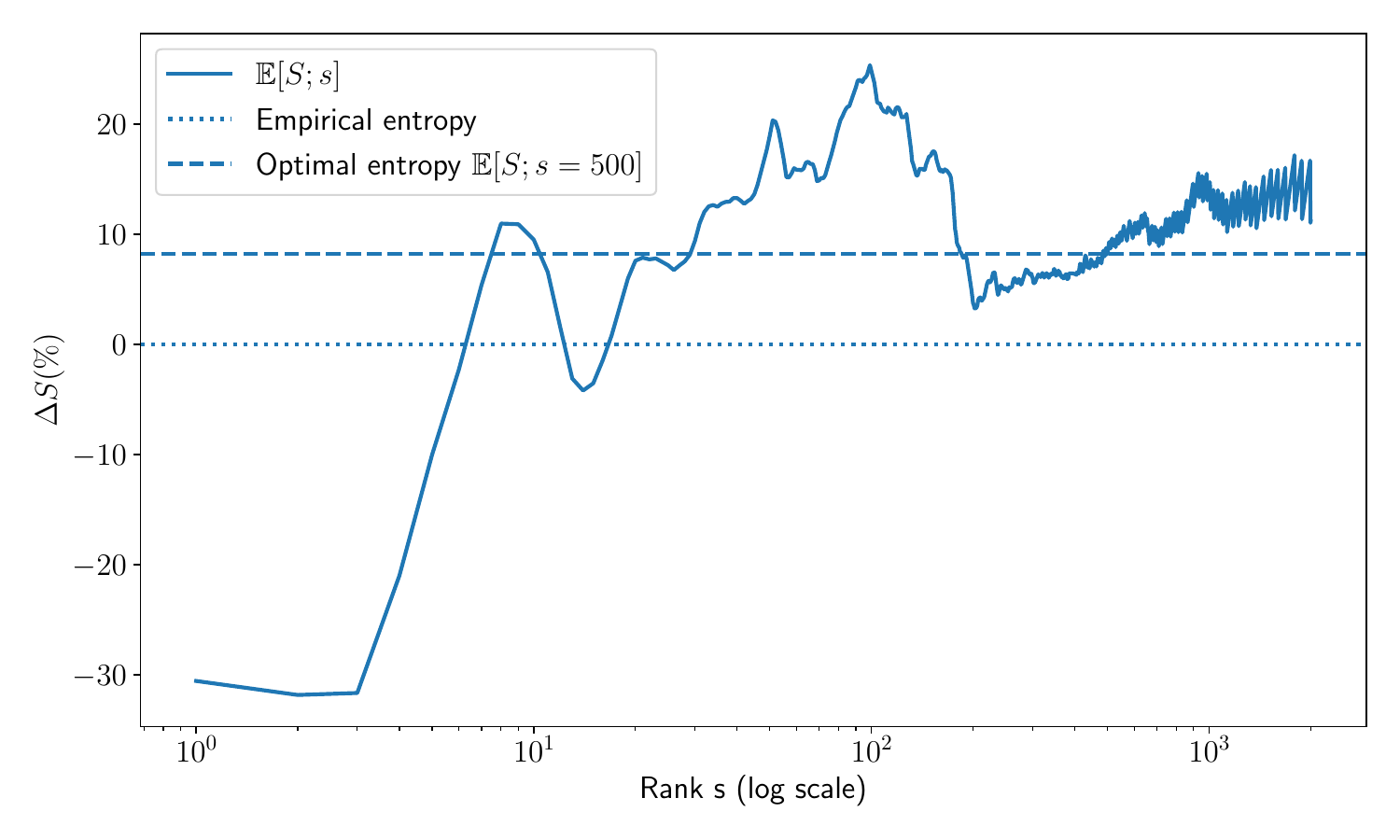}

        \caption{Deviation from empirical entropy of predicted entropy as a function of token rank.}
        \label{fig:dep_s_20ng_func}
    \end{subfigure}
    \hfill
    \begin{subfigure}{0.45\linewidth}
        \includegraphics[width=\linewidth]{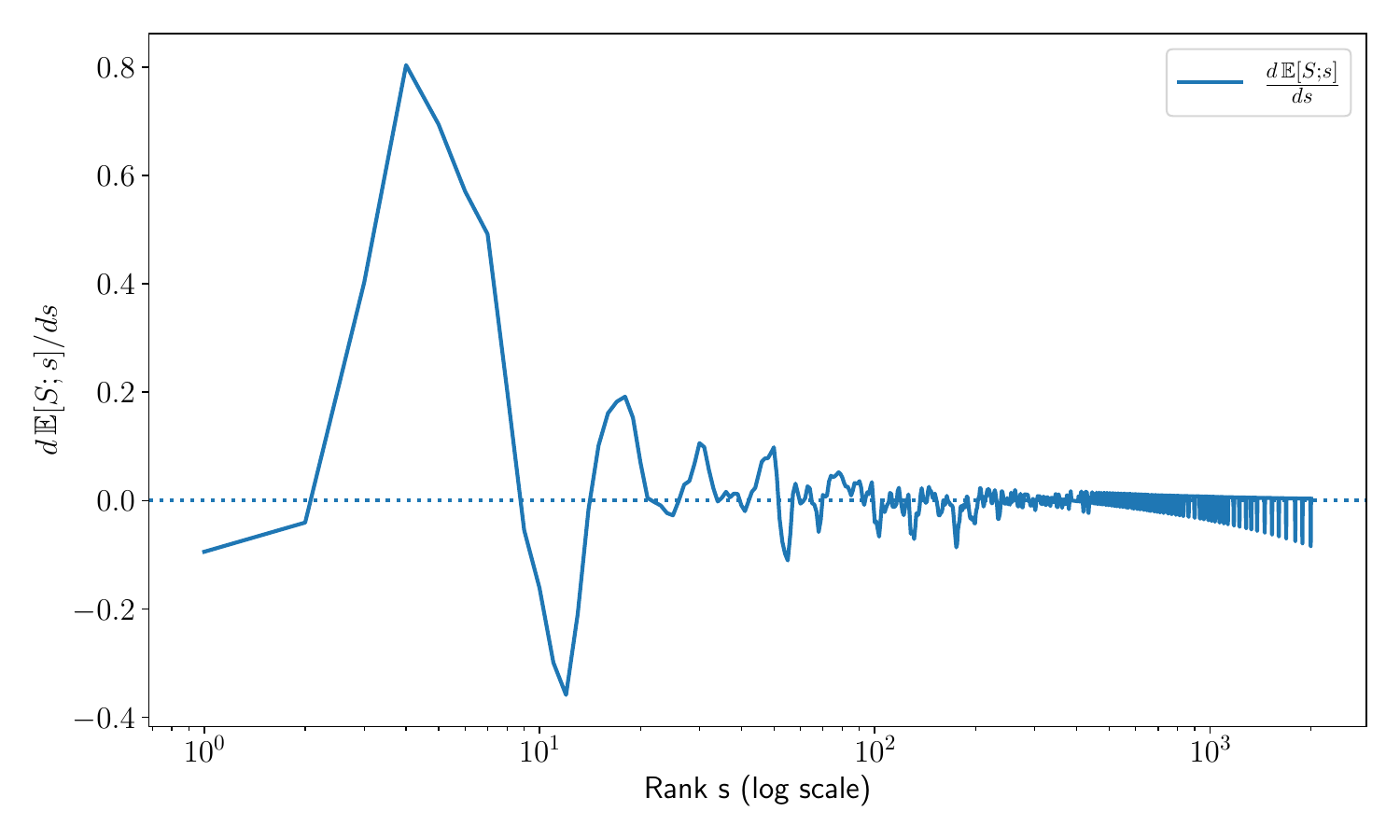}
        \caption{Gradient of predicted entropy as a function of token rank.}
        \label{fig:dep_first_s_20ng_grad}
    \end{subfigure}
    % \begin{subfigure}{0.45\linewidth}
    %     \includegraphics[width=\linewidth]{20Newsgroups_entropy_gradient_linear_zoom_1e2_2e3.pdf}
    %     \caption{Enhanced gradient of predicted entropy as a function of token rank between $10^2$ and $2 \cdot 10^3$.}
    %     \label{fig:dep_first_s_20ng_zoom_grad}
    % \end{subfigure}
    \caption{Dependence of the empirical entropy calculation on $\text{rank}(s)$, the position of the fixed frequency for the 20Newsgroups dataset.}
    \label{fig:dep_s_20ng}
\end{figure}

In Figure \ref{fig:dep_s_20ng}, we present the dependence of the predicted entropy for the most Zipfian corpus (see Figure \ref{fig:zipf_exponents_four_datasets}), namely the 20Newsgroups datasets. Figure \ref{fig:dep_s_20ng_func} shows exactly what is to be expected, away from the tails of the frequency table where the Zipf's law holds strongest,  \eqref{eqn:avg_entopy_zipf_general} is a sufficiently close approximation (to within around 10\%). We excise the rightmost tail where \eqref{eqn:avg_entopy_zipf_general} breaks down for the sake of readability. We note that the noise behaviour in both plots is due to natural fluctuations in $ f_s$ -- some noise has been suppressed with a moving Gaussian average filter\footnote{A stride width of 10 was used in the Gaussian filter.}.

Given this discussion, there is a natural question as to the optimal choice of which rank, $s$ to use to construct the predicted Zipfian entropy. Ideally it should be chosen so the predicted entropy is the stable i.e. where the first and second derivatives of the predicted entropy with respect to $s$ vanish. Thus we will seek to choose $s$ from the region where the gradient is consistently both small and stable. In Figure \ref{fig:dep_first_s_20ng_grad} we plot the gradient of the predicted entropy vs the rank of $s$ ($r_s$); by inspection, one observes  the region where the gradient is both consistently small and stable is between $10^2$ and $2 \cdot 10^3$. In our data, the gradient is most stable around $s=500$, where the predicted entropy $\mathbb E[S; s=500] = 8.05$.

\newpage

\section{Emergence of Zipf's law under byte-pair encoding (BPE)}

Since natural languages seem to almost universally obey Zipf's law, a natural question is whether various encoding schemes used in transformers also behave accordingly? In this exposition, we consider the ubiquitous byte-pair encoding scheme of tokenisation.

It is helpful to measure the fitted Zipf exponent $\alpha(\{f_s\})$ across the corpus, where $\alpha(\{f_s\})$ is defined as the exponent of the line of best fit of the ranks and their relative frequency obtained by straightforward ordinary least squares (OLS) regression.

Since BPE constructs a vocabulary by iteratively merging $n$-gram sequences based on frequency, it is extremely likely that the emergent vocabulary will not correspond to linguistic morphemes or pre-existing words. Deeming whether a Zipfian distribution is emergent in BPE token frequencies will offer insight into the extent to which the encoding scheme preserves, distorts, or rediscovers the linguistic structure of natural language. Such an analysis is particularly relevant for understanding how tokenisation choices influence model behaviour: deviations from Zipfian patterns may shape how semantic information is allocated across tokens. Since natural language empirically obeys the power law behaviour, it is natural to desire it across an encoding. In other words, the persistence of Zipf’s law under BPE could suggest that the algorithm inherits deeper statistical properties of language, and enforces its suitability as a robust tokenisation step for the modelling of natural languages \cite{lotz2025textcompressionevaluatingtokenizers}.

\subsection{Numerical results}

\subsubsection{Recursive applications of BPE lead to Zipf}
\label{ssub:recursive_applications_lead_to_zipf}
\begin{figure}[h]
\centering
\begin{subfigure}[b]{0.45\linewidth}
\centering
\includegraphics[width=\textwidth]{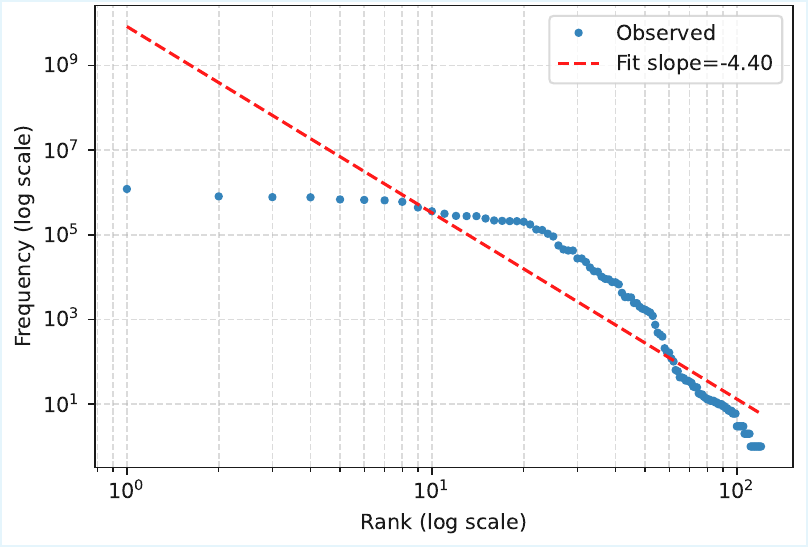}
\caption{Character level}
\end{subfigure}
\begin{subfigure}[b]{0.45\linewidth}
\centering
\includegraphics[width=\textwidth]{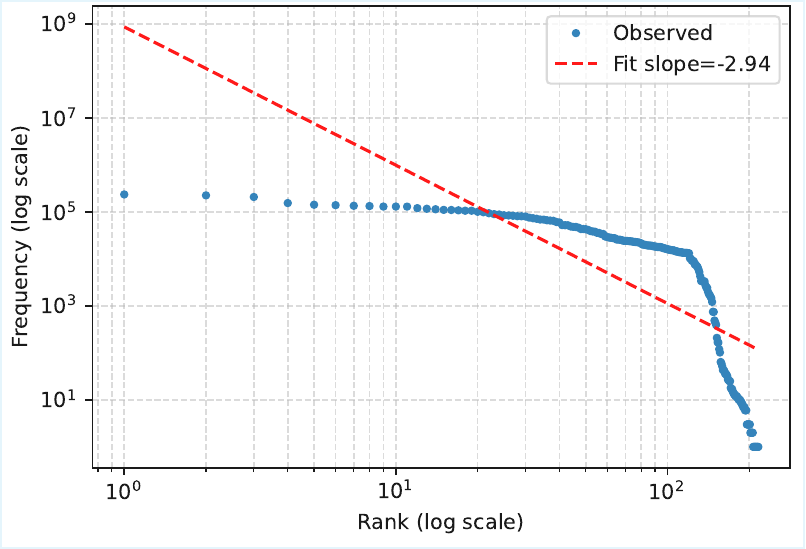}
\caption{100 recursions}
\end{subfigure}\\
\begin{subfigure}[b]{0.45\linewidth}
\centering
\includegraphics[width=\textwidth]{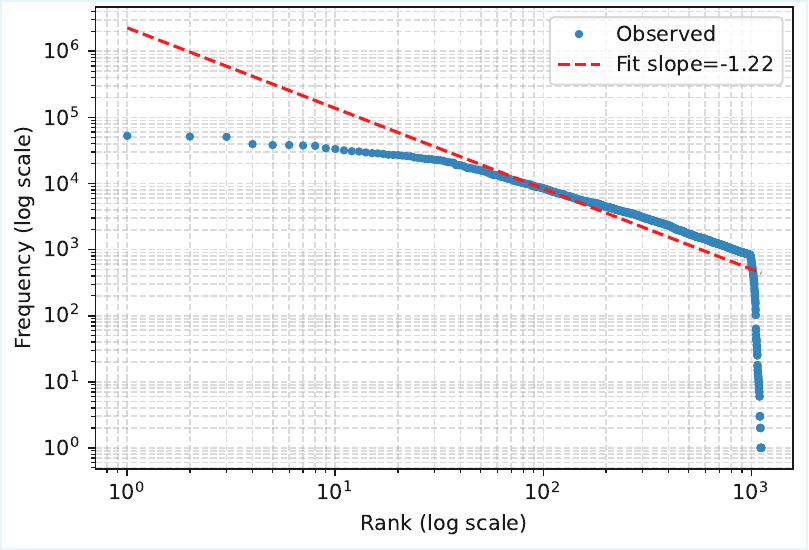}
\caption{1000 recursions}
\end{subfigure}
\begin{subfigure}[b]{0.45\linewidth}
\centering
\includegraphics[width=\textwidth]{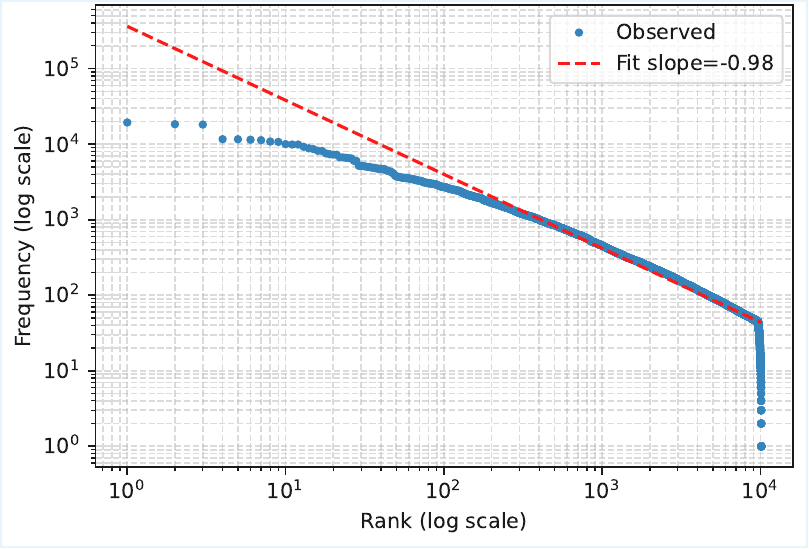}
\caption{10000 recursions}
\end{subfigure}
\caption{Recursive applications of BPE on the IMDB review corpus lead to a corpus which is increasingly Zipfian.}
\label{fig:recursive_bpe_4_stages}
\end{figure}

In this section, we examine the deviation from the power law predicted by Zipf after various iterations $\iota$, and merges $m$ of the BPE algorithm. In Figure \ref{fig:recursive_bpe_4_stages}, we present the fitted power law exponent $\alpha(\{f_s\})$ at various levels of encoding -- this result is in very good agreement with \cite{he2025pretrainedmodelsperformbest} which examined an equivalent procedure on the BookCorpus dataset \cite{bandy2021addressingdocumentationdebtmachine}. Starting with simple character-wise encoding, before extending to many $n$-grams, we observe that the frequencies of tokens become increasingly compliant with Zipf's law as increasing iterations of BPE are applied. This result is predicted in part by the seminal work of \cite{165464}. It is known \cite{165464} that in the regime where each character is equally probable, and for random texts, $\alpha$ is given by
\begin{equation}
\alpha = \frac{\ln(V+1)}{\ln V},
\end{equation}
meaning that as the vocabulary size increases, $\ln (V+1) / \ln V$ asymptotically approaches unity. This argument is incomplete, however. This main contribution relies on the fact that the tokens are IID. Increasing the number of merges create tokens which are much less frequent, more $n$-grams approach the IID (Poisson-like) limit\footnote{See, for example Section \ref{sec:validity_of_iid}.}.

\begin{figure}[h]
    \centering
    \includegraphics[width=0.5\linewidth]{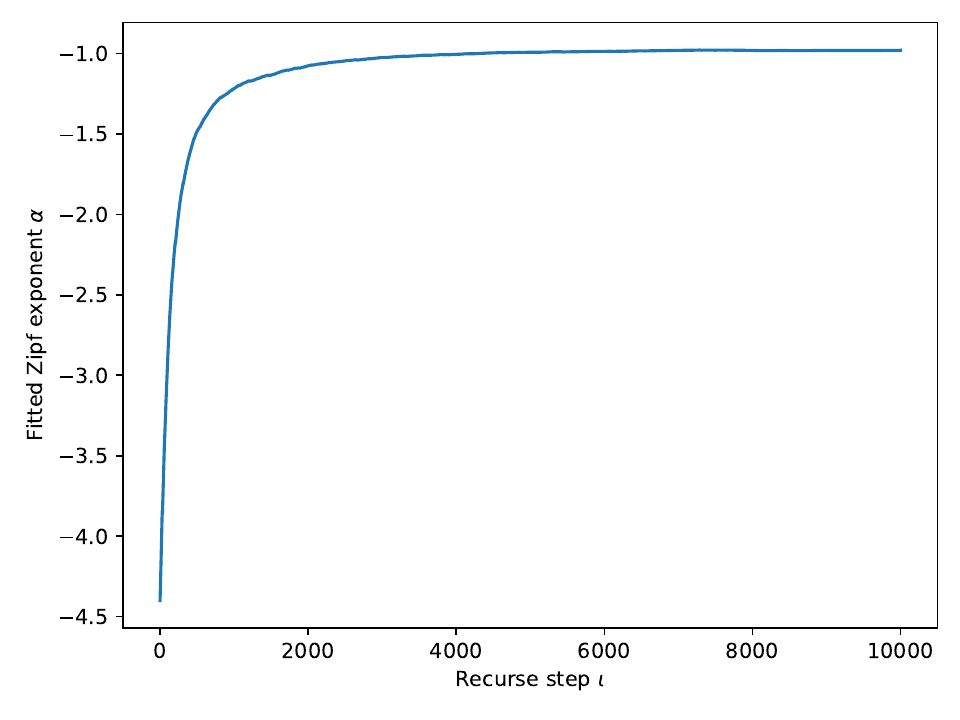}
    \caption{Fitted Zipf exponent under recursive BPE applications for the IMDB dataset.}
    \label{fig:fitted_alpha_bpe_recurse}
\end{figure}

Figure \ref{fig:fitted_alpha_bpe_recurse} shows a set of fits of $\alpha$ across varying numbers of iterations $\iota$, where $\alpha$ is the usual exponent governing the relationship between the frequencies and ranks of tokens in the evolving corpus. At the end of the BPE procedure, almost all tokens lay on the Zipf line, where $\alpha=1$.

\clearpage

\subsection{Empirical entropy under BPE}

Previous discussions have shown how entropy can be a useful measure of randomness and information content in a set of messages. Since Section \ref{sub:entropy_of_a_zipfian_corpus} presented a derivation of entropy in the Zipfian limit of a dataset, for example in the large iteration limit of a corpus undergoing repeated BPE applications, it is a natural exploration to consider both the theoretical and empirical entropies of datasets in this regime. In this section, we consider the empirical entropy of a dataset under recursive applications of byte-pair encoding with a static merges-per-recurse parameter.

For each BPE iteration $\iota$, the corpus may be regarded as an evolving dataset $\mathcal D_\iota = \{m^\mu_\iota\}_{\mu=1}^M$ of $M$ messages, where $m^\mu$ contains 
$N^\mu_\iota$ realised $n$-grams drawn IID from an underlying distribution over a vocabulary 
of size $V$. The composition of the messages is constructed by the algorithm defined in Section \ref{sub:encoding_in_transformers}. Let $n_{s,\iota} := \sum_{\mu=1}^M n_{s,\iota}^\mu$ denote the total number of occurrences 
of type $s$ across the entire dataset, and let 
\[
\Gamma_\iota:= \sum_{\mu=1}^M N^\mu_\iota
\]
denote the total number of $n$-gram slots in $\mathcal D$.  
In the large--$\Gamma_\iota$ limit, the empirical probability of observing a token of type $s$ is 
approximated by
\[
p_{s, \iota} \;=\; \frac{n_{s,\iota}}{\Gamma_\iota},
\]
as in~\eqref{eqn:empirical_prob}.  
Suppressing message and positional indices (which carry no additional information under the IID 
assumption), the empirical average slot entropy is then given by
\begin{equation}
\label{eqn:empirical_entropy_restate}
(S_\iota)_* \;=\; -\sum_{s=1}^V p_{s,\iota} \ln p_{s,\iota}
\;=\; -\sum_{s=1}^V \frac{n_{s,\iota}}{\Gamma_\iota}\,\ln\!\left(\frac{n_{s,\iota}}{\Gamma_\iota}\right),
\end{equation}
which coincides with  \eqref{eqn:finite_entropy}.  
Thus, for each BPE recursion step, one computes the counts $\{n_{s,\iota}\}$ induced by the 
current tokenisation, evaluates the entropy as defined in  \eqref{eqn:empirical_entropy_restate}, and records the 
resulting empirical entropy $S_*$.

\begin{figure}[h]
    \centering
    \includegraphics[width=0.8\linewidth]{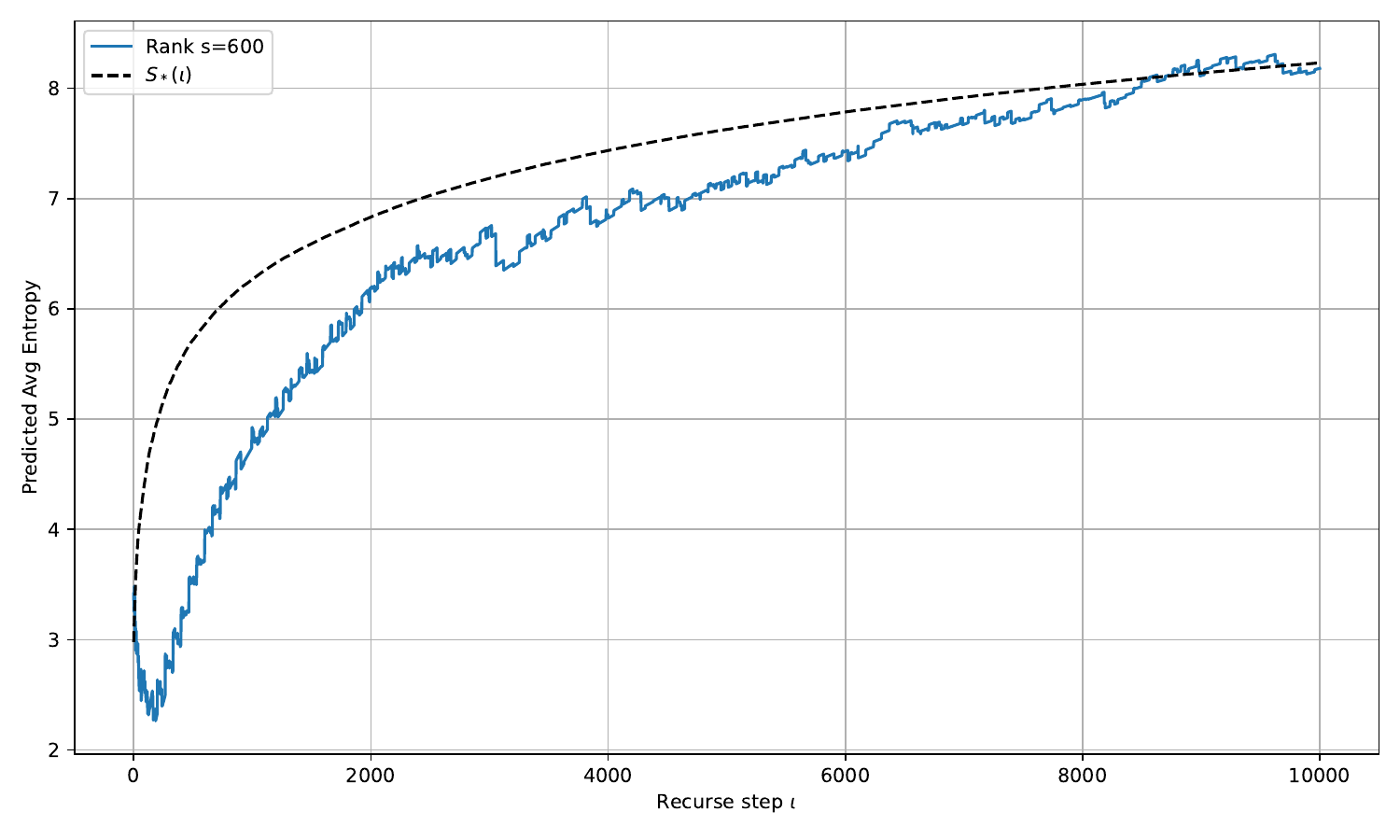}
    \caption{Empirical and predicted entropies as a function of BPE recurse step $\iota$.}
    \label{fig:empirical_corpus_entropy}
\end{figure}

Figure \ref{fig:empirical_corpus_entropy} shows the evolving value of $\mathbb E[S]$ and $S_*$ obtained from  \eqref{eqn:empirical_entropy_restate} as a function of the BPE recursion step $\iota$. For small $\iota$, the vocabulary size $V$ is sufficiently small and the distribution ${n_{s,\iota}}$ contains comparatively few low-frequency types, resulting in a lower measure of entropy. As the merge procedure progresses and the token inventory becomes increasingly fine-grained, the vocabulary expands and many new, infrequent types appear. The empirical entropy therefore rises rapidly for the first several hundred recursion steps. At larger merge depths this growth becomes progressively slower, as additional merges tend to introduce very rare types whose contribution to $-\sum_{s,\iota} p_{s,\iota} \ln p_{s,\iota}$ is small. The resulting curve exhibits a steep initial increase followed by a long, slowly growing tail. 

The empirical behaviour suggests that the entropy of the corpus---as measured by  \eqref{eqn:empirical_entropy_restate}---increases monotonically with tokenisation granularity.
The concave form of the curve indicates diminishing returns: while increasing $\iota$ enlarges the vocabulary and thus broadens the empirical distribution ${p_{s,\iota}}$, the incremental contribution of each newly created rare type becomes smaller at large $V$.
This highlights a trade-off intrinsic to BPE tokenisation: refining the vocabulary increases expressivity and diversity of observed types, but the associated growth in entropy reflects a heavier-tailed empirical distribution and increasing sparsity of counts $n_{s,\iota}$. Furthermore, Figure \ref{fig:empirical_corpus_entropy} shows the predicted entropy using  \eqref{eqn:empirical_entropy_restate} the value of $s$ such that $\text{rank}(s)$ is both in the centre of the frequency table and where the gradient of the expected entropy with respect to changing $s$ is sufficiently small (the gradient here is broadly comparable to Figure \ref{fig:dep_first_s_20ng_grad}, and thus omitted for brevity). As with Section \ref{ssub:dep_of_entropy_on_s}, we once more observe that the entropy prediction breaks down at the tails for sufficiently small or large $f_s$, however holds sufficiently well provided $\text{rank}(s)$ is in the centre of the dataset. 

\section{An exploration with transformers}%
\label{sec:zipfian_transformers}

Previous discussions have centred on empirical measurements solely with large datasets, and independently of models. In this section we explore a GPT-2 transformer \cite{hf_canonical_model_maintainers_2022} trained on the IMDB dataset \cite{imdb_reviews}. Employing a transformer network with a softmax activation function applied to the logits allows a well-defined, correctly normalised, probability distribution across the vocabulary for each position to be constructed. From the softmax, it is possible to define bona-fide entropy independently of any large dataset limits. As such, we present the following experiment: consider a selection of networks trained subject to a text completion objective, each corresponding to differing tokenisation levels. In Section \ref{ssub:recursive_applications_lead_to_zipf} we have empirically observed in large BPE application limit, the Zipf deviation $|\alpha(\{f_s\}) - 1|$ becomes closer to 0. 

\subsection{Numerical setup}
\label{sub:numerical_explanation_with_transformers}
To analyse how tokenization level changes the informational properties of model outputs, it is useful to distinguish between two complementary perspectives on entropy. When a language model generates text, it produces at each step a probability distribution over the vocabulary via a softmax function. Drawing from that distribution dictates which token is ultimately selected. One may therefore characterize the model's behaviour either by examining the entropy of this predictive distribution itself, or by analysing the entropy of the sequence of sampled tokens that results from it. The former captures the model's internal uncertainty before sampling, while the latter reflects the empirical diversity of its realised outputs. One should note that these two notions of entropy, whilst closely related, are not equivalent in small sample limits.

\paragraph{Entropy of the Model's Softmax Distribution.}
To quantify the model's uncertainty at the point of generation, we evaluate the entropy of the softmax distribution at each output step. Given a model parametrisation that produces logits $z
\in \mathbb{R}^V$ over a vocabulary of size $V$, the corresponding token probabilities are obtained via the softmax transformation
\begin{equation}
p_s = \frac{\exp(z_s)}{\sum_{s'=1}^V \exp(z_{s'})}.
\end{equation}
The instantaneous predictive entropy is then defined as
\begin{equation}
H(p) = -\sum_{s=1}^V p_s \log p_s,
\end{equation}
which measures the dispersion of the model's belief over the token space. This entropy reflects the model’s internal uncertainty prior to sampling or decoding and is therefore independent of any specific generation strategy. By computing this quantity across a corpus of model outputs, one can characterize how tokenization granularity affects the sharpness or flattening of the predictive distribution, particularly as the vocabulary grows and BPE merges approach a regime in which token frequencies approximate a Zipfian structure.

\paragraph{Entropy of the Sampled Token Sequence.}
In addition to analysing the model's predictive entropy, one may also examine the empirical entropy of the sampled token sequence produced by the model after the softmax. Let $\{x_t\}_{t=1}^{T}$ denote a generated sequence with empirical token frequencies $f_s$. In the simultaneous IID (used in  \eqref{eqn:IID_entropy}) and large BPE iteration ($\iota$) limits, one expects the distribution will become Zipfian and  \eqref{eqn:avg_entopy_zipf_general} will dominate for sufficiently sensible $s$.

Intuitively, the sample entropy
\begin{equation}
H_* = -\sum_{s \in \mathcal{V}} p_*(s)\log p_*(s)
\end{equation}
captures how the model's realised outputs distribute over the vocabulary. Unlike the softmax-based entropy, which reflects model uncertainty at the distributional level, the sample entropy reflects the realized diversity of the generated text and depends both on the predictive probabilities and on the stochasticity introduced by the sampling procedure. In the large sample, large-vocabulary BPE limit---where the corpus approaches a Zipfian distribution---the relationship between these two notions of entropy becomes informative for understanding how the model allocates probability mass across increasingly fine-grained $n$-gram structures.

\subsubsection{Numerical results}
\label{ssub:softmax_entropy_numerical}

\begin{figure}[h]
\centering
\begin{subfigure}[b]{0.45\linewidth}
\centering
\includegraphics[width=\textwidth]{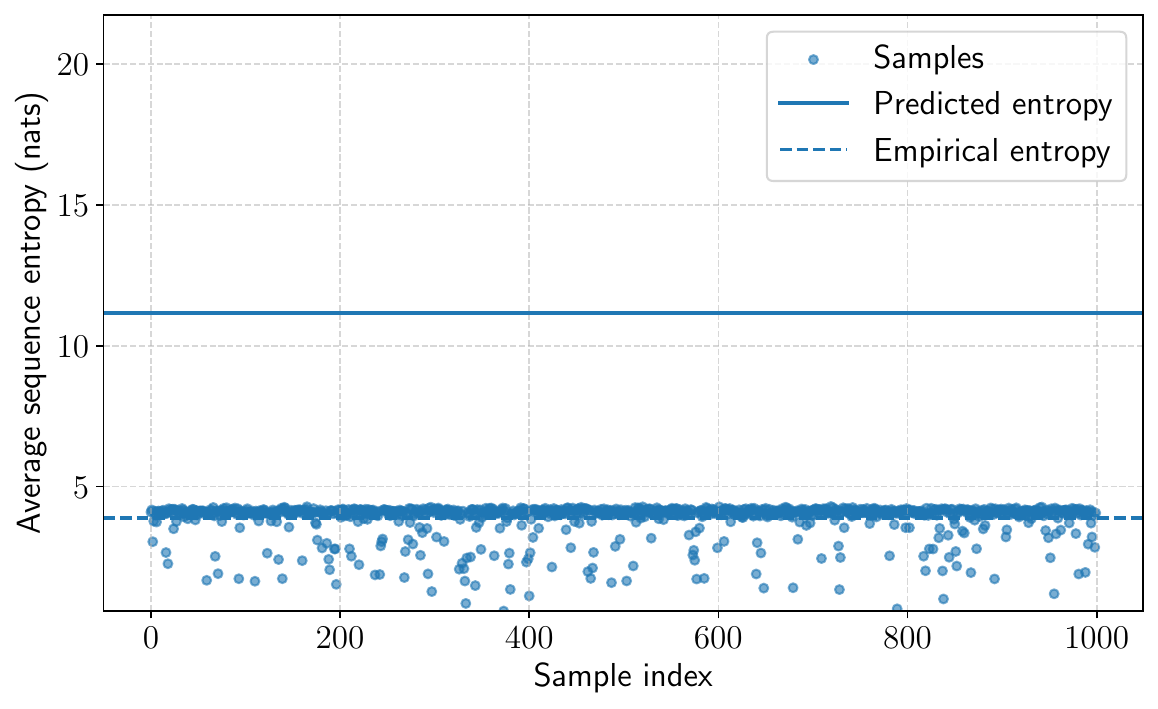}
\caption{100 recurse steps}
\end{subfigure}
\begin{subfigure}[b]{0.45\linewidth}
\centering
\includegraphics[width=\textwidth]{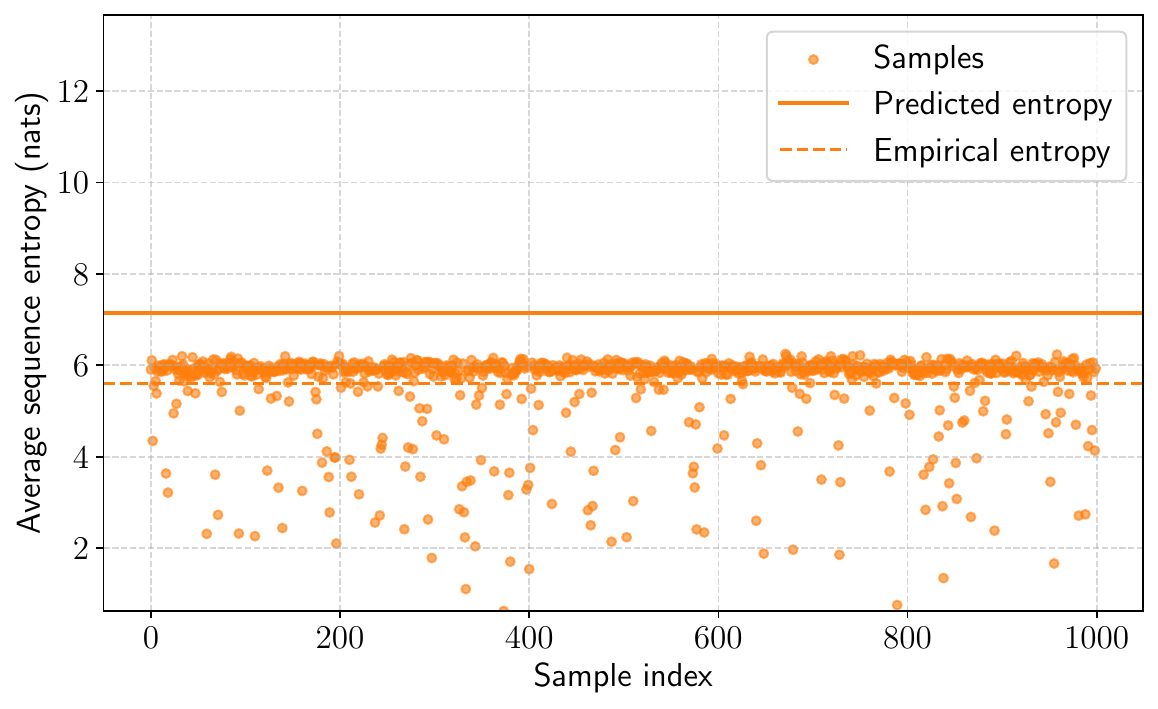}
\caption{1000 recurse steps}
\end{subfigure}\\
\begin{subfigure}[b]{0.45\linewidth}
\centering
\includegraphics[width=\textwidth]{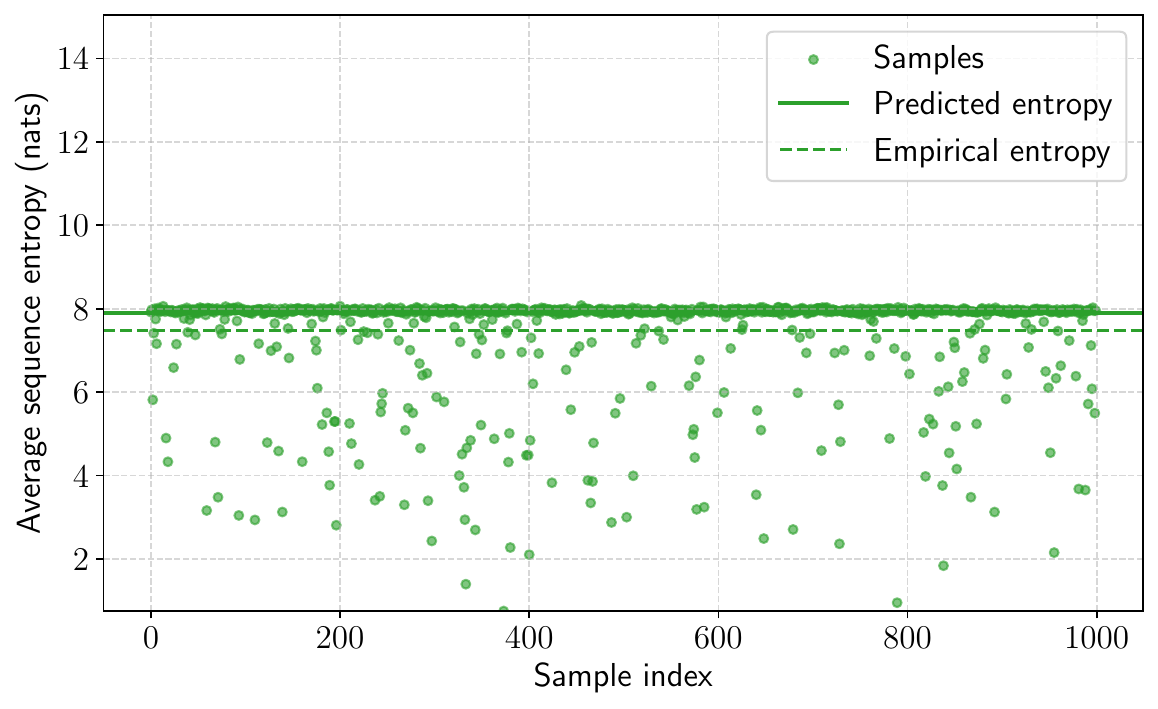}
\caption{10000 recurse steps}
\end{subfigure}
\begin{subfigure}[b]{0.45\linewidth}
\centering
\includegraphics[width=\textwidth]{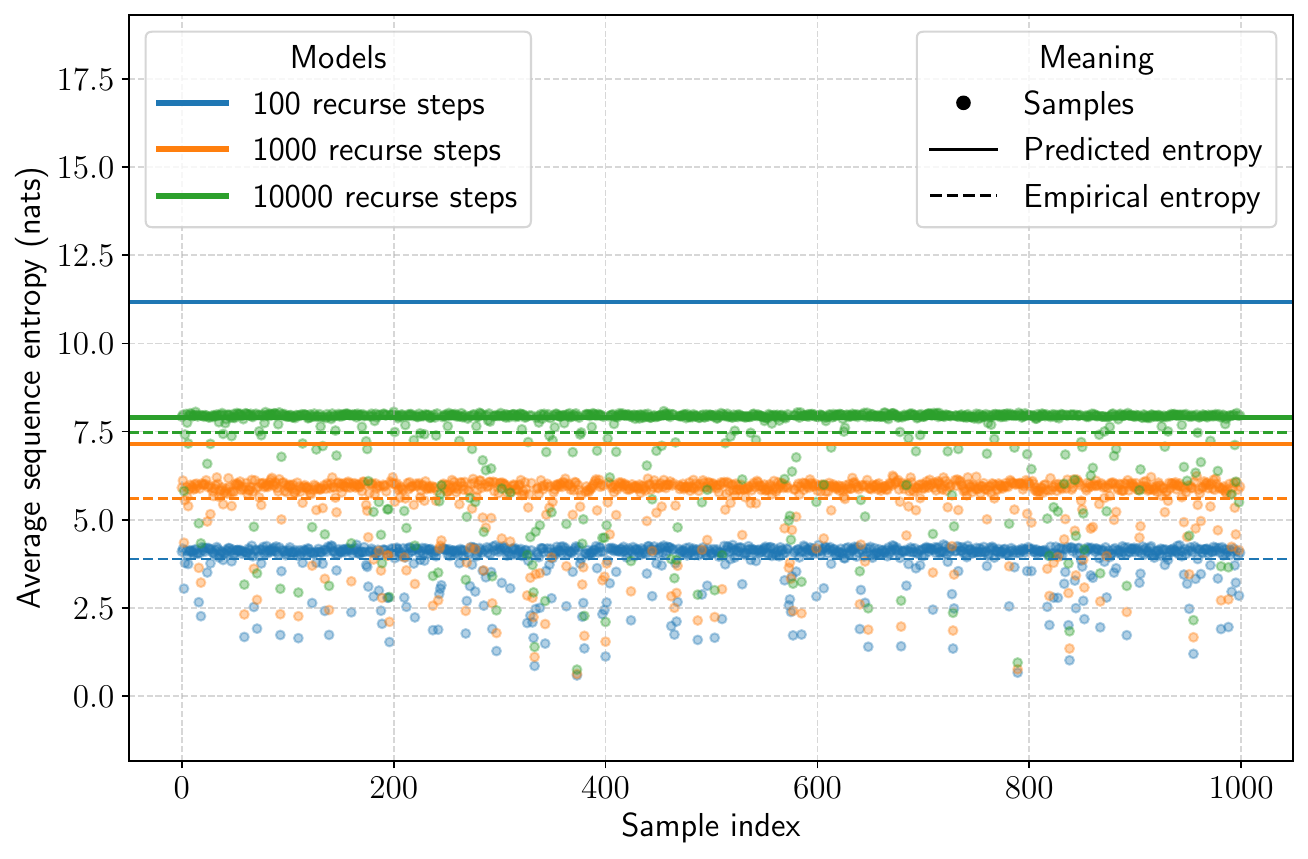}
\caption{Combined}
\label{fig:softmax_empirical_entropy_combined}
\end{subfigure}
\caption{Entropies for increasing merge depths from the network softmax function.}
\label{fig:combined_entropies_softmax}
\end{figure}

Figure \ref{fig:combined_entropies_softmax} shows the empirical and predicted average entropies for the transformer outputs conditioned on messages from the IMDB corpus\footnote{As has been seen, one must make a  choice of $s$ in Equation \eqref{eqn:avg_entopy_zipf_general}. Here we choose $s=1$ safely as the tails of the frequencies at the output of the transformer more closely obey Zipf's law.}.
Previous experiments, e.g. Section \ref{ssub:recursive_applications_lead_to_zipf}, have shown that the models with the larger vocabulary, i.e. more encoding applications, more closely follow Zipf's law. This is corroborated vicariously through the entropies of the transformer network. The plots depict the empirical mean entropy by dashed lines, and the predicted mean entropy obtained through the IID assumption in solid lines. Around the character level (100 recurse steps),  \eqref{eqn:avg_entopy_zipf} overestimates the entropy by approximately three times. This is a result of a pair of factors: in order to utilise the aforementioned equation, each tokens must be IID -- this is a poorly saturated limit at such a recursion depth (see Section \ref{sec:validity_of_iid}). Secondly and more obviously, for small merge depths, the transformer output poorly obeys Zipf's law. For this dataset, the predicted entropy approaches the empirical value from above, agreeing remarkably strongly in the large recursion depth limit. Figure \ref{fig:combined_entropies_softmax} combines each plot, and exemplifies this behaviour more strongly.

\clearpage
\subsection{Validity of the IID assumption}
\label{sec:validity_of_iid}
In Section \ref{ssub:softmax_entropy_numerical}, the discussion relied on the fact that $n$-grams were independent and identically distributed. Given the transformer model, one may probe violations of the IID assumption as follows. Consider an arbitrary head in a \textit{sufficiently well-trained} transformer-based large language model\footnote{In this work all transformers are `well-trained' where losses have converged etc.}.

In the attention mechanism \cite{vaswani2023attentionneed}, the off-diagonal key-query interactions can be interpreted as a measure of statistical dependence  between $n$-grams in the input corpus. Let $Q \in \mathbb{R}^{T \times d}$ and $K \in \mathbb{R}^{T \times d}$  denote the query and key matrices for a sequence of length $T$, and let $A$ be the un-normalised attention score matrix as in \cite{vaswani2023attentionneed}. The diagonal entries $A_{ii}$ capture how strongly position $i$ attends to itself, while the off--diagonal terms  $A_{ij}$ for $i \neq j$ represent interactions between distinct $n$-grams.

If the corpus exhibits strong cross-token dependencies, then for many $i$ there exist $j \neq i$ such that $A_{ij}$ is large, indicating
that the query at position $i$ finds its nearest neighbours (in the key space) at other positions. In contrast, if the $n$-grams are largely independent, the matrix $A$ will be approximately diagonal, namely
\begin{equation}
|A_{ij}| \ll |A_{ii}| \quad \text{for most } i \neq j,
\end{equation}
signalling that each $n$-gram primarily interacts with itself or its local 
context.  

Therefore, the distribution of off--diagonal magnitudes in $A$ provides a 
natural quantitative proxy for the independence of $n$-grams: larger and more 
structured off--diagonal values imply dependence, while small or diffuse 
off--diagonal values indicate relative independence across the corpus. 

In practice, modern language models possess large numbers of attention heads, meaning this procedure must be repeated many times. It is customary to introduce a neighbour distance cutoff to minimise the computation complexity. This is a realistic assumption, in natural language; if a pair of tokens are very far from one another temporally, they are significantly less likely to be meaningfully correlated. Given the canonical (temporal) ordering of the tokens, one may create a set of average local attention values for each token in the message, namely
\begin{equation}
D_i := \{a_{ij} \;|\; |j-i|\leq \Lambda\},
\end{equation}
for a given cutoff $\Lambda$. The distribution of $\{D_i\}$ then acts as a proxy for the neighbour correlation for each $n$-gram.

\begin{figure}
    \centering
    \includegraphics[width=0.8\linewidth]{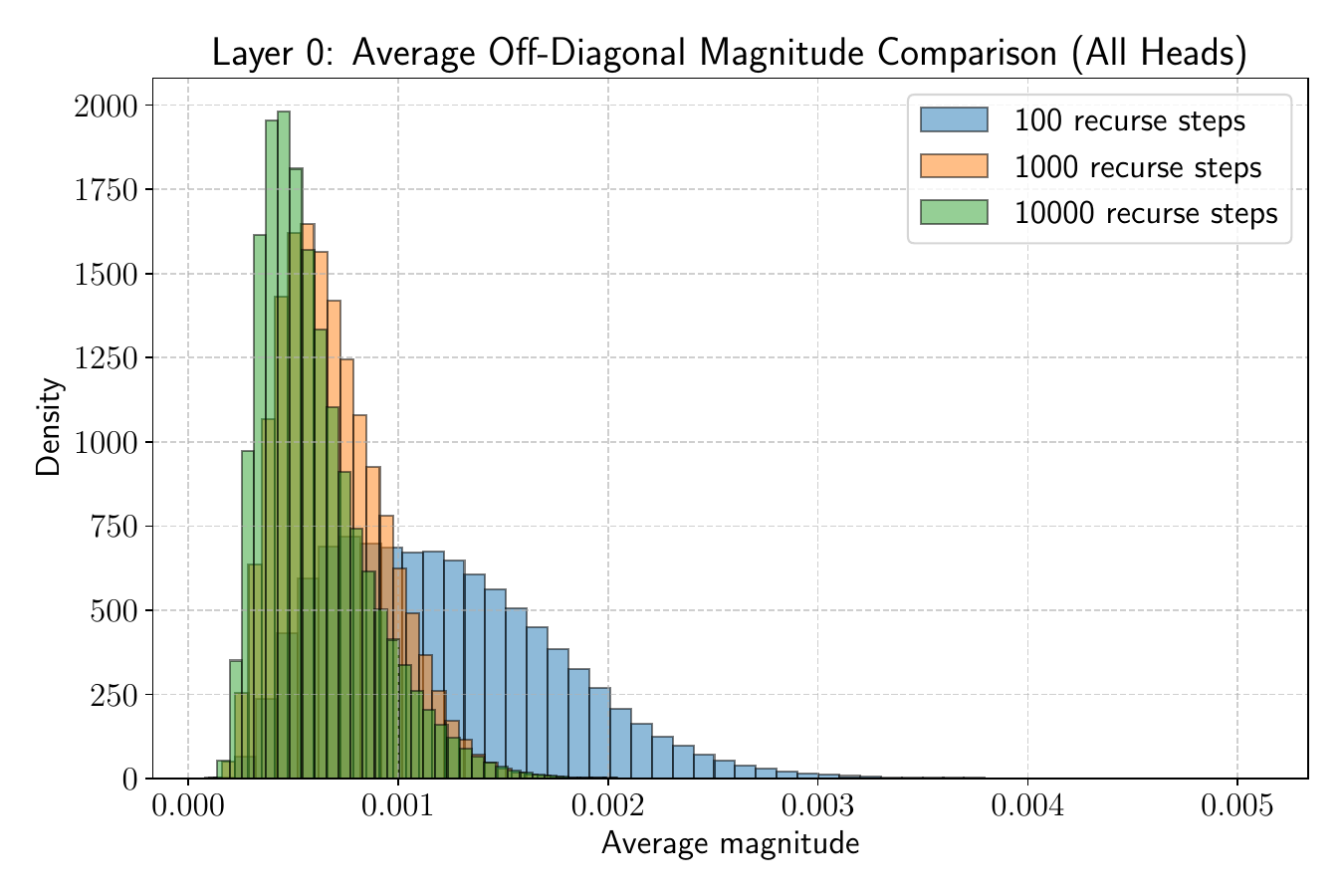}
    \caption{Magnitude of the local normalised off-diagonal elements of the attention matrix.}
    \label{fig:histograms_attention_values}
\end{figure}

Since the GPT-2 model used in this experiment contains many attention heads, one must make a choice which to examine. Figure \ref{fig:histograms_attention_values} shows histograms of pooled $A_{ij}$ for all heads in the model. The same behaviour is manifest in each attention head independently (see Appendix \ref{sub:numerical_explanation_with_transformers} for specifics on head numbers in the model). As can be seen, increasing the number of encoding recurses $\iota$ results in a corpus where the coupling between sufficiently local $n$-grams, as measured by the attention matrix, decreases. The mean of the distribution of attention values becomes increasingly close to 0 for increasing applications of BPE. Moreover, the variance of the distribution becomes smaller meaning the number of neighbouring tokens which `correlate' also decreases, on average.

\clearpage

\section{Conclusion}

In this work we have examined the statistical structure of corpora subjected to byte–pair encoding and its relation to the emergence of Zipfian behaviour, with particular emphasis on the associated entropy of the resulting token distributions. Beginning from a theoretical analysis of Shannon entropy under an exactly Zipfian frequency law, we derived an explicit expression for the average slot entropy that depends only on the size of the vocabulary and the highest-frequency token. This establishes a clear quantitative prediction for the informational content of any corpus whose token distribution sufficiently approximates the law.

Empirically, we demonstrated that recursive applications of BPE consistently drive token frequencies toward Zipf’s law across a range of corpus granularities as controlled by the number of BPE applications. As the number of merges increases, the fitted Zipf exponent converges toward unity, and the empirical entropy rises sharply before entering a regime of diminishing returns. This behaviour reflects the expansion of the vocabulary into increasingly fine-grained but increasingly rare subword units. The resulting entropy curves closely mirror the theoretical expectations derived in the Zipfian limit, with discrepancies at small merge depths attributable to both non-Zipfian frequency structure and the failure of the IID assumption in the early stages of tokenisation.

By extending the analysis to transformer models trained on the same corpora, we further observed that the entropic properties of a model's softmax outputs track the Zipfian structure of the underlying tokenisation. Models trained with larger vocabularies---corresponding to larger BPE recursion depths---exhibit predictive entropies that more closely match the Zipf-predicted values, indicating that the model internalises and reproduces the statistical regularities of the corpus. Complementary attention-based diagnostics suggest that deeper tokenisation reduces local token–token correlations, rendering the empirical distribution increasingly compatible with the independence assumptions underlying the theoretical entropy calculations.

Taken together, these results indicate that byte–pair encoding not only compresses the linguistic structure into a workable subword vocabulary, but also progressively reconstructs the core statistical regularities of natural language as reflected in Zipf's law. The emergence of predictable entropic behaviour under BPE provides a principled foundation for understanding how tokenisation choices shape the informational landscape encountered by transformer models. More broadly, our findings highlight entropy as a unifying tool for characterising both corpora and the models trained on them, offering a bridge between empirical linguistic structure and the probabilistic mechanics of modern neural language models.

Finally, we re-emphasise the analogy between renormalisation and linguistic tokenisation. In terms of the language of renormalisation group flows, the asymptotics of Byte Pair Encoding is the linguistic equivalent of a flow to the infrared. The Zipf's law statistics manifested in that limit then appear as the linguistic infrared universality class.

\section*{Data availability statement}

The code for this project is available on GitHub at \href{https://github.com/xand-stapleton/natural-language-tokenisation}{https://github.com/xand-stapleton/natural-language-tokenisation}. All scripts, algorithms, and links to resources necessary for replication of the results are included within the repository.

\section*{Acknowledgements}
\label{sec:acknowledgements}

DSB and AGS acknowledge support from Pierre Andurand over the course of this research, and DSB is also supported by the Science and Technology Facilities Council (STFC) Consolidated Grant ST/T000686/1 ``Amplitudes, strings \& duality.''  AGS wishes thank J. Kerrison for his comments about relevant research on the relationship between tokenisation and the encoding of protein structures in SOTA models.

% ---------------
% End of main body
% ---------------
\newpage
\bibliographystyle{unsrt}
\bibliography{bib}
\newpage
\appendix

\section{Description of transformer task}
\label{app:transformer_specs}
This work used the \texttt{GPT2LMHeadModel} from HuggingFace \cite{hf_canonical_model_maintainers_2022}, fine-tuned on the IMDB movie reviews dataset for sentiment-aware text generation. The model is based on OpenAI's GPT-2 architecture and employs multiple transformer decoder layers, each with multi-head self-attention and feed-forward networks. Table~\ref{tab:gpt2lmheadmodel} summarizes the key architectural parameters of this model variant for the 10k recurse step tokenisation.

This work used the model in its pre-trained and fine-tuned form. Each transformer block contains a multi-head self-attention mechanism, which allows the model to capture long-range dependencies between tokens, and a feed-forward network that processes these representations. We restrict the context window to 128 tokens to balance computational efficiency with sufficient local context for sentence- and paragraph-level review completion. Table \ref{tab:gpt2lmheadmodel} outlines the technical specification of the model used for the large vocabulary; for smaller vocabularies, the parameters were identical, except the embedding layers which were scaled accordingly.

% \begin{table}[h!]
% \centering
% \begin{tabular}{l c}
% \hline
% \textbf{Parameter} & \textbf{Value} \\
% \hline
% Number of layers (transformer blocks) & 12 \\
% Hidden size (dimensionality of embeddings) & 768 \\
% Number of attention heads & 12 \\
% Context window (maximum sequence length) & 1024 tokens \\
% Vocabulary size & 50257 \\
% Total parameters & 124M \\
% Feed-forward hidden size & 3072 \\
% Activation function & GELU \\
% Fine-tuned corpus & IMDB movie reviews \\
% \hline
% \end{tabular}
% \caption{Key architecture parameters of the IMDB-fine-tuned GPT2LMHeadModel used in this work.}
% \label{tab:gpt2-params}
% \end{table}
\begin{table}[h]
\centering
\small
\begin{tabular}{ll}
\toprule
\textbf{Component} & \textbf{Specification} \\
\midrule
Model & \texttt{GPT2LMHeadModel} \\
Transformer backbone & \texttt{GPT2Model} \\

Word token embedding (wte) & Embedding $(10115 \times 256)$ \\
Positional embedding (wpe) & Embedding $(128 \times 256)$ \\
Embedding dropout & Dropout $(p = 0.1)$ \\

Number of transformer blocks & $6 \times$ \texttt{GPT2Block} \\

LayerNorm (ln\_1) & LayerNorm $(256,\ \epsilon = 10^{-5})$ \\
Self-attention module & \texttt{GPT2Attention} \\
Attention projection (c\_attn) & Conv1D $(768 \rightarrow 256)$ \\
Output projection (c\_proj) & Conv1D $(256 \rightarrow 256)$ \\
Attention dropout & Dropout $(p = 0.1)$ \\
Residual dropout & Dropout $(p = 0.1)$ \\

LayerNorm (ln\_2) & LayerNorm $(256,\ \epsilon = 10^{-5})$ \\
Feed-forward network & \texttt{GPT2MLP} \\
FC layer (c\_fc) & Conv1D $(1024 \rightarrow 256)$ \\
Projection layer (c\_proj) & Conv1D $(256 \rightarrow 1024)$ \\
Activation function & NewGELU \\
MLP dropout & Dropout $(p = 0.1)$ \\

Final LayerNorm (ln\_f) & LayerNorm $(256,\ \epsilon = 10^{-5})$ \\
Language model head & Linear $(256 \rightarrow 10115)$ \\
\bottomrule
\end{tabular}

\caption{Architecture and hyperparameters of the GPT2LMHeadModel used in this work.}
\label{tab:gpt2lmheadmodel}
\end{table}

\section{Datasets and context dependence}
In all experiments contained within this note, we use one of four widely used benchmark datasets for text classification.
\begin{itemize}
    \item The \textit{20 Newsgroups} \cite{Newsgroups20} dataset consists of approximately 20,000 documents evenly distributed across 20 thematic newsgroups, covering topics such as politics, sports, religion, and science.
20Newsgroups is commonly used for document classification and context analysis tasks.

\item The \textit{IMDB Movie Review} dataset \cite{imdb_reviews} contains approximately 50,000 English-language movie reviews labelled with binary sentiment (positive or negative), with an equal split between training and test sets. In this work, notwithstanding training and evaluation of the GPT2 model, we combine test and training datasets into a monolithic corpus.

This dataset is a standard benchmark for sentiment analysis and focuses on long-form, opinionated text.

\item The \textit{AG News} dataset \cite{gulli_corpus_2023} AGNews is a large-scale news classification corpus composed of over 120,000 news articles categorized into four classes: World, Sports, Business, and Science/Technology.
It is frequently used for evaluating topic classification models on short to medium-length texts.

\item The \textit{SMS Spam Collection} dataset \cite{sms_spam_collection_228} comprises approximately 5500 SMS messages labelled as either spam or ham (non-spam).
Due to its short message length and class imbalance, it is commonly employed to study binary text classification and robustness to noisy, informal language.
\end{itemize}

Table \ref{tab:dataset_stats} also presents a series of key dataset characteristics.

\begin{table}[h]
\centering
\caption{Summary statistics of the text classification datasets used in our experiments.}
\label{tab:dataset_stats}
\begin{tabular}{lcccc}
\hline
\textbf{Dataset} & \textbf{\#Samples} & \textbf{\#Classes} & \textbf{Avg. Length} & \textbf{Task} \\
\hline
20 Newsgroups & $\sim$20,000 & 20 & $\sim$250 words & Topic classification \\
IMDB Reviews & 50,000 & 2 & $\sim$230 words & Sentiment analysis \\
AG News & 120,000 & 4 & $\sim$40 words & News classification \\
SMS Spam & 5,574 & 2 & $\sim$15 words & Spam detection \\
\hline
\end{tabular}
\end{table}

\end{document}